\documentclass[letterpaper, 10 pt, conference]{ieeeconf}
\IEEEoverridecommandlockouts
\usepackage{amsmath,amssymb}
\usepackage{graphicx}
\usepackage{booktabs}
\usepackage{multirow}
\usepackage{soul}
\usepackage{xcolor}
\usepackage[normalem]{ulem}
\usepackage{tikz}
\usepackage{booktabs}
\usepackage{tabularx}
\usepackage{array}
\usetikzlibrary{arrows.meta,positioning,calc,fit,shapes.geometric}

\newcommand{\pmv}[2]{#1{\scriptsize$\pm$#2}}

\newcommand{\sg}{\operatorname{sg}}
\newcommand{\LN}{\operatorname{LN}}
\newcommand{\ours}{PRA}
\newcommand{\oursitem}{PRA-item}
\newcommand{\ourscond}{PRA-cond}

\title{\LARGE \bf
Toward Cross-Rubric Surgical Skill Assessment via Joint Embedding Prediction
}

\author{Jiajun Cheng$^{1}$,
Haidar M. Abdul-Muhsin$^{2}$,
Juha Choi$^{1}$, and
Shan Lin$^{1}$%
\thanks{$^{1}$Jiajun Cheng, Juha Choi, and Shan Lin are with the
School of Electrical, Computer and Energy Engineering,
Arizona State University, Tempe, AZ, USA.
{\ttfamily\small \{ccheng58, jchoi230, shan.lin.2\}@asu.edu}}%
\thanks{$^{2}$Haidar M. Abdul-Muhsin is with the
Department of Urology, Mayo Clinic, Phoenix, AZ, USA.
{\ttfamily\small abdul-muhsin.haidar@mayo.edu}}%
}

\begin{document}

\maketitle
\thispagestyle{empty}
\pagestyle{empty}



\begin{abstract}
Expert review of surgical videos is essential for surgical training but labor-intensive, motivating
AI-assisted skill assessment to provide feedback at scale.
Although surgical assessment rubrics may use different language to describe similar skills, existing automated methods commonly predict predefined skill levels
or numerical scores tied to fixed rubrics, limiting their ability to transfer across semantically equivalent assessment criteria.
We propose Predictive Rubric Alignment (PRA), a joint
embedding prediction approach that scores surgical videos
using natural-language descriptions of skill levels.
Rather than matching level probabilities, PRA predicts a
rater-weighted combination of skill-level description
embeddings.
Adjacent description features define a score-parameterized path,
and the predicted embedding is projected onto this path to read the score.
We introduce two variants: \textit{PRA-item}, which uses learned queries for predefined assessment items, and \textit{PRA-cond}, which generates queries from textual item names to enable assessment of unseen items without retraining.
On AIxSuture, a dataset of expert-rated suturing videos,
PRA outperforms published baselines in skill classification.
We further evaluate zero-shot transfer to the smaller surgical skill assessment
benchmark JIGSAWS.
Trained on only two AIxSuture rubric items, PRA-cond
assesses unseen JIGSAWS items without further training, yielding
positive score correlations and greater Grad-CAM
overlap with instruments and suture than a
description-matching baseline.
These results support predictive rubric alignment as a promising
direction for flexible and explainable automated surgical skill
assessment.
\end{abstract}

\section{Introduction}

Surgical skill assessment plays a critical role in surgical training and quality
improvement~\cite{birkmeyer2013}. In today's hospitals, attending surgeons still need to dedicate substantial time and effort to reviewing surgical videos and evaluating trainees' performance.
This labor-intensive process limits the scalability of surgical training, motivating AI-assisted skill assessment at
scale~\cite{igaki2023}.
Existing methods typically learn from expert ratings based on standardized rubrics, such as the Objective Structured Assessment of
Technical Skills (OSATS) ~\cite{martin1997} (Table~\ref{tab:rubric_items}), by predicting
predefined skill levels~\cite{funke2019,hoffmann2024} or regressing numerical
scores ~\cite{anastasiou2023}. 
While effective within their training settings, these models are tied to fixed assessment criteria. 
As surgical competencies and their descriptions vary across procedures and rubrics~\cite{larson2005,toale2023}, this dependence limits their flexibility in assessing skills beyond the criteria used during training.

Vision--language models (VLMs) align visual inputs with free-form
text, supporting zero-shot recognition of concepts specified
through new text prompts~\cite{radford2021,zhai2023siglip}.
This generalization ability motivates exploring skill assessment
with text-defined criteria and performance levels, allowing
transfer to new rubric items and video settings without further
training. Related work spans ordinal prediction and continuous
visual quality scoring (Sec.~II). In surgery, VLM-based skill
assessment remains underexplored, and prompting general models
to score performance remains challenging~\cite{rau2025}.
\begin{figure}[t]
\centering
\resizebox{\columnwidth}{!}{\input{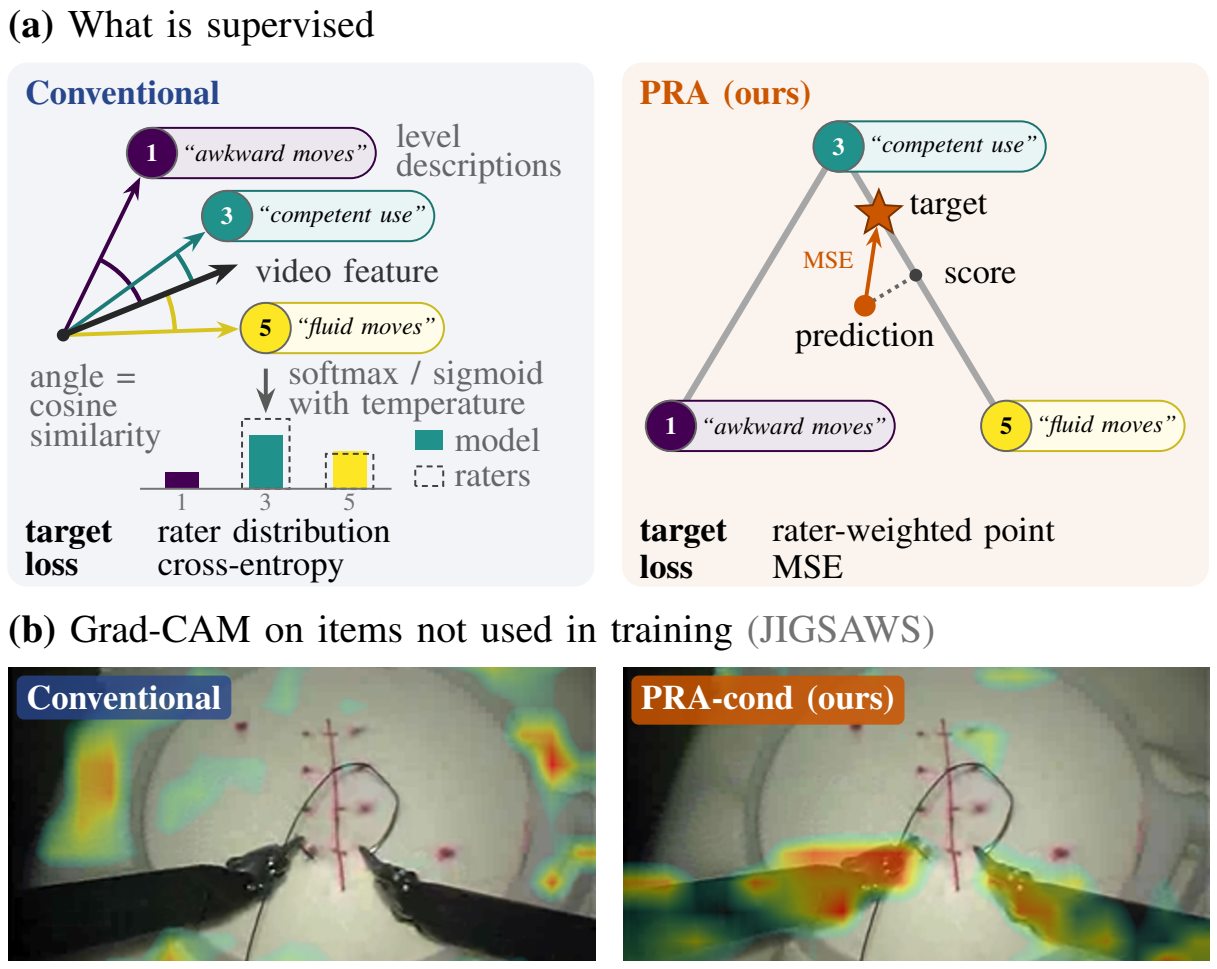}}
\caption{(a)~Conventional methods train level probabilities from the cosine similarities between the video feature and
the level descriptions, with (binary) cross-entropy; \ours{} instead regresses one point between the descriptions in latent space.
(b)~Grad-CAM on rubric items not used in training shows our method focuses on the relevant visual evidence region compared with the conventional method (binary cross entropy).}

\label{fig:teaser}
\end{figure}

\begin{table}[t]
\centering
\caption{Rubric items and performance-level descriptions.}
\label{tab:rubric_items}
\footnotesize
\setlength{\tabcolsep}{4pt}
\renewcommand{\arraystretch}{1.12}

\begin{tabularx}{\columnwidth}{
    @{}>{\raggedright\arraybackslash}X
    >{\raggedright\arraybackslash}X@{}
}
\toprule
\textbf{AIxSuture item} & \textbf{JIGSAWS item} \\
\midrule
Respect for tissue
& Respect for tissue \\

Time and motion
& Time and motion \\

Instrument handling
& -- \\

Suture technique
& Suture/needle handling \\

Flow of operation \& forward planning
& Flow of operation \\

Knowledge of specific procedure
& -- \\

Quality of final product
& Quality of final product \\

Overall performance
& Overall performance \\
\bottomrule
\end{tabularx}

\par\vspace{3pt}
\begin{minipage}{\columnwidth}
JIGSAWS items use the OSATS level descriptions associated
with the same row; -- indicates no corresponding item.
\end{minipage}

\par\vspace{5pt}

\begin{tabularx}{\columnwidth}{
    @{}c >{\raggedright\arraybackslash}X@{}
}
\toprule
\textbf{Score}
& \textbf{Instrument handling: level descriptions} \\
\midrule
1
& Repeatedly makes tentative or awkward moves with instruments. \\

3
& Competent use of instruments, although occasionally appeared
stiff or awkward. \\

5
& Fluid moves with instruments and no awkwardness. \\
\bottomrule
\end{tabularx}
\end{table}
Adapting these VLM-based approaches to surgical skill
assessment requires considering both how scores are computed
and what visual evidence supports them.
Conventional contrastive learning often converts cosine similarities into level probabilities and
introduces a temperature scale through softmax or sigmoid
(Fig.~\ref{fig:teaser}(a)), whose choice or initialization can
affect learning and downstream
performance~\cite{agarwala2023temperature,zhai2023siglip}. Additionally, 
Global image--text contrastive alignment also does not guarantee
that the vision encoder preserves fine-grained visual
details~\cite{mukhoti2023pacl,dong2023maskclip}.
This concern is particularly relevant when rubric descriptions
express high-level performance rather than detailed video content,
potentially encouraging reliance on contextual shortcuts rather
than evidence relevant to the
rating~\cite{cheng2026surgxbench,cheng2026trajpred,cheng2026lavif}.
These considerations motivate exploring an alternative to
probability matching while examining the visual evidence
supporting its scores.

We propose Predictive Rubric Alignment (PRA), which predicts
a location in the rubric's description embedding space rather
than probabilities over its levels
(Fig.~\ref{fig:teaser}(a)).
Each performance-level description is encoded as a vector.
For each rubric item, we connect adjacent description vectors
to form a path parameterized by score.
Rater scores determine a weighted combination of these vectors,
which the video model learns to predict by minimizing squared
distance to this target.
At inference, we find the closest point on the path and return
its corresponding score.
Following LeVLJEPA~\cite{kuhn2026}, we use a joint embedding
predictive architecture (JEPA) with bidirectional video--text
embedding prediction rather than contrastive matching.
LeVLJEPA also reports stronger spatial features than
contrastive learning.
We develop two versions of PRA for fixed and text-specified rubrics.
PRA-item assesses the rubric items used during training,
whereas PRA-cond additionally accepts item names as text to
specify what to assess.
New rubric items and their performance descriptions can therefore
be supplied to PRA-cond without further training.

Our contributions are threefold.
\begin{enumerate}
\item We propose PRA, which formulates surgical skill scoring
as predicting a rater-weighted combination of performance-level
description embeddings and reads the score by projection onto
the description path, providing an alternative to
level-probability matching.

\item We introduce PRA-cond, which accepts rubric item names
and performance descriptions as text, enabling zero-shot
assessment of rubric items not used during training.

\item We evaluate PRA on AIxSuture and zero-shot transfer
to JIGSAWS, demonstrating improved AIxSuture performance
over published baselines and same-encoder controls,
positive zero-shot score correlations, and greater Grad-CAM
overlap with instruments and suture than a
description-matching baseline.
\end{enumerate}

\section{Related Work}
\paragraph{Surgical skill and action quality assessment}
Video-based surgical skill assessment has been formulated as
classification into predefined skill levels on JIGSAWS and
AIxSuture~\cite{funke2019,hoffmann2024}, and as regression of
numerical skill scores~\cite{anastasiou2023}. To account for
ambiguity in ratings, Tang et al.~\cite{tang2020} train models
to predict score distributions, with evaluations on sports
and surgical videos. Language has also been introduced to guide
the visual features used for scoring: Xu et al.~\cite{xu2024}
use CLIP to align action terminology with video-, stage-, and
frame-level representations, evaluating on benchmarks including
JIGSAWS.

\paragraph{Vision--language ordinal prediction and scoring}
Building on CLIP~\cite{radford2021}, OrdinalCLIP interpolates rank features into ordered language prototypes~\cite{li2022ordinalclip}, while L2RCLIP adds RankFormer and a cross-modal ordinal pairwise loss~\cite{wang2023l2rclip}. CLIP-based quality scorers include LIQE, which learns through multitask correspondence and quality ranking~\cite{zhang2023liqe}, and CLIP-PCQA, which supervises predicted quality distributions using EMD and quantile losses~\cite{liu2025clippcqa}. Autoregressive approaches include Q-Align, which trains on discretized textual levels and reads continuous scores from level-token probabilities~\cite{wu2024qalign}, and DeQA-Score, which matches these probabilities to score-distribution soft labels using KL divergence~\cite{you2025deqa}. PRA instead regresses a rater-weighted combination of rubric features and reads its position relative to the description path. We implement this formulation with LeVLJEPA's bidirectional cross-modal prediction and stop-gradient targets~\cite{kuhn2026}, together with LeJEPA's distributional regularization~\cite{balestriero2025}.

\section{METHOD}
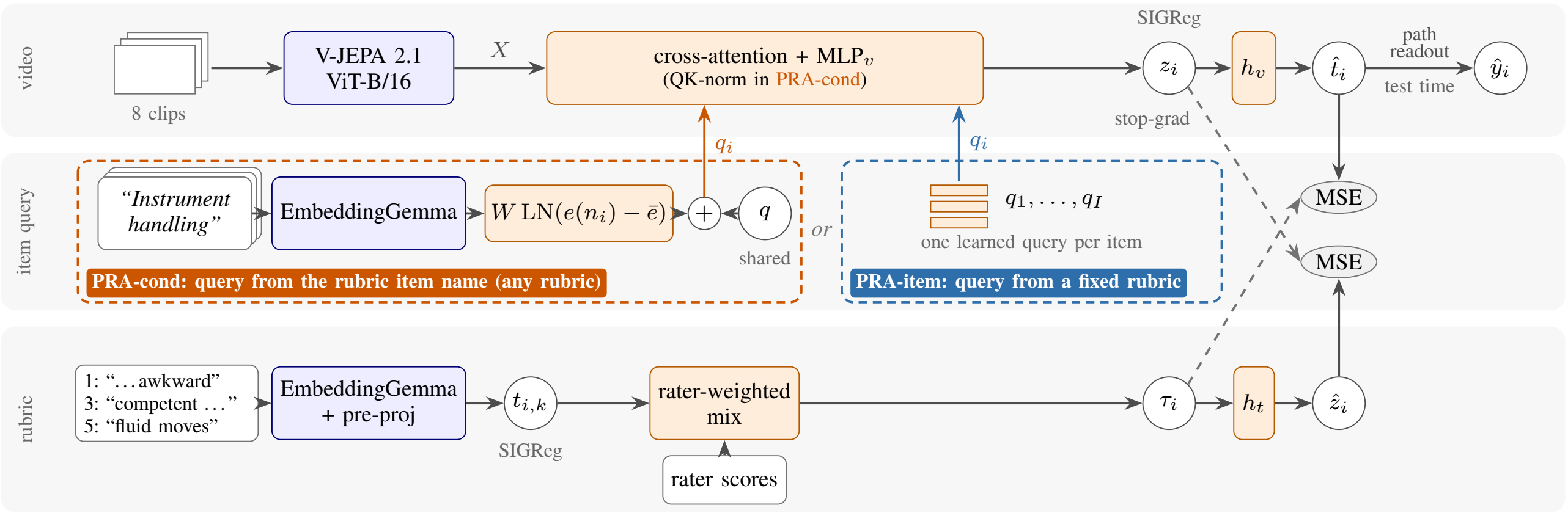
\begin{figure*}[t]
\centering
\resizebox{0.97\textwidth}{!}{
\definecolor{condc}{RGB}{204,85,0}
\definecolor{itemc}{RGB}{46,110,170}
\begin{tikzpicture}[
  box/.style={draw=black!55, rounded corners=3pt, align=center, font=\footnotesize, minimum height=9mm, inner sep=3pt,
              fill=white},
  enc/.style={box, fill=blue!7, draw=blue!45!black},
  ours/.style={box, fill=orange!14, draw=orange!70!black},
  vec/.style={draw=black!70, circle, inner sep=1.3pt, font=\footnotesize, fill=white, minimum size=6.5mm},
  arr/.style={-{Stealth[length=2.2mm]}, thick, black!70},
  sgarr/.style={-{Stealth[length=2.2mm]}, thick, black!55, dashed},
  note/.style={font=\scriptsize, text=black!60, align=center},
  variant/.style={draw=#1, densely dashed, rounded corners=4pt, line width=0.8pt},
  vtag/.style={font=\scriptsize\bfseries, text=white, fill=#1, rounded corners=1.5pt, inner sep=2pt,
               anchor=south west}]
\fill[black!3, rounded corners=5pt] (-0.95,0.8) rectangle (18.4,-0.85);
\fill[black!3, rounded corners=5pt] (-0.95,-1.05) rectangle (18.4,-3.0);
\fill[black!3, rounded corners=5pt] (-0.95,-3.2) rectangle (18.4,-5.5);
\node[note, rotate=90] at (-0.65,0) {video};
\node[note, rotate=90] at (-0.65,-2.02) {item query};
\node[note, rotate=90] at (-0.65,-4.35) {rubric};
\foreach \d in {0.16,0.08,0} {\draw[fill=white, draw=black!55] (0.45+\d,-0.32+\d) rectangle (1.45+\d,0.28+\d);}
\node[note] at (1.0,-0.58) {8 clips};
\node[enc, minimum width=2.1cm] (venc) at (3.6,0) {V-JEPA 2.1\\ViT-B/16};
\node[ours, minimum width=5.4cm] (pool) at (8.5,0) {cross-attention + MLP$_v$\\[-1pt]{\scriptsize(QK-norm in \textcolor{condc}{\ourscond{}})}};
\node[vec] (z) at (13.5,0) {$z_i$};
\node[ours] (hv) at (14.55,0) {$h_v$};
\node[vec] (that) at (15.6,0) {$\hat t_i$};
\node[vec] (yhat) at (17.6,0) {$\hat y_i$};
\draw[arr] (1.65,0) -- (venc); \draw[arr] (venc) -- node[above, font=\footnotesize] {$X$} (pool);
\draw[arr] (pool) -- (z); \draw[arr] (z) -- (hv); \draw[arr] (hv) -- (that);
\draw[arr] (that) -- node[above, font=\scriptsize, align=center] {path\\[-2pt]readout} node[below, note] {test time} (yhat);
\node[note, above=0.4mm of z] {SIGReg};
\draw[variant=condc] (0.0,-1.15) rectangle (8.95,-2.9);
\node[vtag=condc] at (0.1,-2.85) {\ourscond{}: query from the rubric item name (any rubric)};
\foreach \d in {0.12,0.06} {\draw[fill=white, draw=black!55, rounded corners=3pt]
  (0.25+\d,-2.25+\d) rectangle (2.15+\d,-1.35+\d);}
\node[box, font=\footnotesize\itshape, minimum width=1.9cm] (name) at (1.2,-1.8) {``Instrument\\handling''};
\node[enc, minimum width=2.1cm] (nenc) at (3.6,-1.8) {EmbeddingGemma};
\node[ours, minimum height=7mm, inner xsep=2pt] (shift) at (6.2,-1.8) {$W\,\mathrm{LN}(e(n_i)-\bar e)$};
\node[circle, draw=black!70, fill=white, inner sep=0pt, minimum size=4mm, font=\footnotesize] (plus) at (7.75,-1.8) {$+$};
\node[vec] (qs) at (8.5,-1.8) {$q$};
\node[note, anchor=north] at (8.5,-2.14) {shared};
\draw[arr] (name) -- (nenc); \draw[arr] (nenc) -- (shift); \draw[arr] (shift) -- (plus); \draw[arr] (qs) -- (plus);
\draw[arr, condc] (plus) -- node[right, pos=0.55, font=\footnotesize] {$q_i$} (plus |- pool.south);
\node[font=\footnotesize\itshape, text=black!55] at (9.2,-2.02) {or};
\draw[variant=itemc] (9.45,-1.15) rectangle (14.15,-2.9);
\node[vtag=itemc] at (9.55,-2.85) {\oursitem{}: query from a fixed rubric};
\foreach \r in {0,1,2} {\draw[fill=orange!14, draw=orange!70!black] (10.55,-1.45-0.2*\r) rectangle (11.25,-1.58-0.2*\r);}
\node[font=\footnotesize, anchor=west] at (11.35,-1.65) {$q_1,\dots,q_I$};
\node[note] at (11.8,-2.17) {one learned query per item};
\draw[arr, itemc] (10.9,-1.4) -- node[right, pos=0.5, font=\footnotesize] {$q_i$} (10.9,-1.4 |- pool.south);
\node[box, align=left, font=\scriptsize, text width=2.05cm] (desc) at (1.1,-4.15)
  {1: ``\dots awkward''\\3: ``competent \dots''\\5: ``fluid moves''};
\node[enc, minimum width=2.1cm] (tenc) at (3.6,-4.15) {EmbeddingGemma\\+ pre-proj};
\node[vec] (temb) at (5.6,-4.15) {$t_{i,k}$};
\node[note, below=0.4mm of temb] {SIGReg};
\node[ours] (mix) at (8.0,-4.15) {rater-weighted\\mix};
\node[box, minimum height=6mm] (raters) at (8.0,-5.1) {rater scores};
\node[vec] (tau) at (13.5,-4.15) {$\tau_i$};
\node[ours] (ht) at (14.55,-4.15) {$h_t$};
\node[vec] (zhat) at (15.6,-4.15) {$\hat z_i$};
\draw[arr] (desc) -- (tenc); \draw[arr] (tenc) -- (temb); \draw[arr] (temb) -- (mix); \draw[arr] (raters) -- (mix);
\draw[arr] (mix) -- (tau); \draw[arr] (tau) -- (ht); \draw[arr] (ht) -- (zhat);
\node[draw=black!55, ellipse, fill=black!6, font=\footnotesize, inner sep=1.2pt] (msev) at (15.6,-1.6) {MSE};
\node[draw=black!55, ellipse, fill=black!6, font=\footnotesize, inner sep=1.2pt] (mset) at (15.6,-2.4) {MSE};
\draw[arr] (that) -- (msev); \draw[arr] (zhat) -- (mset);
\draw[sgarr] (tau.north east) -- (msev.west);
\draw[sgarr] (z.south east) -- (mset.west);
\node[note, anchor=east] at (13.88,-0.64) {stop-grad};
\end{tikzpicture}}
\caption{PRA architecture. Item queries attend to V-JEPA~2.1
video tokens, producing item vectors $z_i$. PRA-cond derives
queries from item names; PRA-item learns one query per item.
Predictors $h_v$ and $h_t$ regress stop-gradient targets
$\tau_i$ and $z_i$ in opposite directions. The rater-weighted
target $\tau_i$ combines level-description features, and SIGReg
regularizes video and description features. At inference,
path projection converts $h_v(z_i)$ into a score.}
\label{fig:pipeline}
\end{figure*}
\subsection{Problem setup}
A rubric consists of $I$ items, each an aspect of the performance with a name $n_i$ (e.g.\ ``Instrument handling'')
and written descriptions of some of its score levels. AIxSuture uses the $I{=}8$ OSATS items~\cite{martin1997}
(Table~\ref{tab:rubric_items}), which describe the levels $\mathcal{K}=\{1,3,5\}$, so item $i$ comes with three level
descriptions, each encoded as ``\emph{item name}: \emph{level description}''. A video is rated on every item by $R$ raters ($R{=}3$) with
integer scores $y_{i,r}\in\{1,\dots,5\}$; the label of item $i$ is the raters' mean $\bar y_i$, and 80\% of the
AIxSuture labels lie at no described level. The model predicts item scores $\hat y_i\in[1,5]$; the predicted global rating score (GRS) is
$\sum_i\hat y_i$ and the rater GRS is $\sum_i\bar y_i$, and the skill class follows with the official thresholds~\cite{hoffmann2024} (novice $\mathrm{GRS}<16$, intermediate
$16\leq\mathrm{GRS}<24$, proficient $\mathrm{GRS}\geq24$)

\subsection{Encoders and item vectors}
A video is split into 8 equal segments with one 16-frame clip (4\,fps) each. V-JEPA~2.1 ViT-B/16~\cite{vjepa21}
encodes each clip at $224\times384$ into 2688 tokens; with a learnable clip-position feature, the $8\times2688$
tokens form the set $X$. Level descriptions and item names are encoded by EmbeddingGemma-300m~\cite{featuregemma} with mean
pooling, $e(\cdot)\in\mathbb{R}^{768}$, and a pre-projection MLP maps the description of level $k$ to its description feature
$t_{i,k}\in\mathbb{R}^{d}$, $d{=}768$. Both encoders are fine-tuned. Item $i$ reads the video with its own query $q_i$
through a cross-attention block,
\begin{equation}
\bar x_i=\mathrm{Attn}(q_i, X),\qquad z_i=\mathrm{MLP}_v(\bar x_i)\in\mathbb{R}^{d},
\end{equation}
so a rubric with $I$ items gives $I$ queries and $I$ item vectors (8 on AIxSuture). We propose two versions of \ours{}
that differ only in where the queries come from. In \oursitem{}, the $I$ queries are learned directly, one vector per
item, so the model is tied to the items it was trained on. In \ourscond{}, they are computed from one shared learnable
query and the item names (Sec.~\ref{sec:cond}), so the model has no item-specific parameters and can be given a new
rubric as text.

\subsection{Rubric-based target construction}
\label{sec:stops}
A rater score is mapped to weights on the described levels by linear interpolation, averaged over the raters,
\begin{equation}
\phi_k(y)=\big[1-\tfrac{|y-k|}{2}\big]_+,\qquad w_i=\tfrac1R\textstyle\sum_{r}\phi(y_{i,r}),
\label{eq:weights}
\end{equation}
so that $\sum_k k\,w_{i,k}=\bar y_i$; when some raters score below 3 and others above it, e.g.\ 2, 3 and 4, $w_i$ is
spread over all three levels and keeps their disagreement. \ours{} mixes the description features with these weights,
\begin{equation}
\tau_i=\textstyle\sum_{k\in\mathcal{K}}w_{i,k}\,t_{i,k},
\label{eq:target}
\end{equation}
The three description features of item $i$ define an affine plane. Since the rater weights are nonnegative and sum to one, the target $\tau_i$ lies within the triangle formed by these features. The
squared distance to $\tau_i$ needs \textbf{no scale or temperature}.

\subsection{Predictive rubric alignment}
\label{sec:pra}
A squared error in the raw feature space with both encoders trained collapses without a counter-term. Following
LeVLJEPA~\cite{kuhn2026}, a video-to-text predictor $h_v$ and a text-to-video predictor $h_t$ regress the
stop-gradient feature of the other modality, and SIGReg~\cite{balestriero2025} keeps the video vectors (per
item query, across the videos of the batch on each GPU) and the description features close to an isotropic Gaussian to prevent collapse:
\begin{multline}
\mathcal{L}=\tfrac{1-\lambda_v-\lambda_t}{2}\Big[\big\|h_v(z_i)-\sg(\tau_i)\big\|^2+\big\|h_t(\tau_i)-\sg(z_i)\big\|^2\Big]\\
+\lambda_v\,\mathrm{SIGReg}(\{z_i\})+\lambda_t\,\mathrm{SIGReg}(\{t_{i,k}\}),
\label{eq:loss}
\end{multline}
averaged over the items and videos of a batch ($\|\cdot\|^2$: mean squared error over the $d$ dimensions).

\subsection{Score readout}
\label{sec:readout}
A readout turns the model's output feature for item $i$ into a score $\hat y_i\in[1,5]$; we use two following readout methods.

\textit{Path readout.} The description features of item $i$, connected in score order, form a path
$t_{i,1}\to t_{i,3}\to t_{i,5}$ on which the point of score $y\in[1,5]$ is $p_i(y)=\sum_{k}\phi_k(y)\,t_{i,k}$, with the
weights of~\eqref{eq:weights} (e.g.\ $p_i(2)$ is the midpoint of $t_{i,1}$ and $t_{i,3}$). The path readout returns the
score of the path point closest to the prediction $\hat t_i$,
\begin{equation}
\hat y_i=\operatorname*{arg\,min}_{y\in[1,5]}\big\|\hat t_i-p_i(y)\big\|.
\label{eq:readout}
\end{equation}
For \ours{}, $\hat t_i=h_v(z_i)$. Since the target uses the same weights, $\tau_i=p_i(\bar y_i)$ when $w_i$ falls on two
neighbouring levels, so a prediction that reaches its target is read back as the raters' mean score, without a
temperature or calibration. For BCE and CE-8, which compare vectors by cosine similarity, $\hat t_i$ is the item vector
$z_i$ (BCE: the shared vector $z$), and all vectors are scaled to unit length first.

\textit{Probability readout.} BCE and CE-8 also output a probability $P_{i,k}$ for each described level
(Sec.~\ref{sec:compared}; for BCE, the three sigmoids of item $i$ divided by their sum). The probability readout is the
expected level,
\begin{equation}
\hat y_i=\textstyle\sum_{k}k\,P_{i,k},
\label{eq:probreadout}
\end{equation}
\ours{} outputs a point, not probabilities, and is read only on the path.

\subsection{Rubric-item conditioning}
\label{sec:cond}
\ourscond{} shifts a shared learnable query
$q\in\mathbb{R}^{d}$ using item name $n_i$:
\begin{equation}
q_i=q+W\,\LN\big(e(n_i)-\bar e\big),\qquad
\bar e=\tfrac1I\textstyle\sum_{j=1}^{I}e(n_j).
\label{eq:cond}
\end{equation}
Here $e(n_i)\in\mathbb{R}^{768}$ encodes the item name,
$\bar e$ averages the rubric's item-name features, $\LN$
is layer normalization, and the learnable
$W\in\mathbb{R}^{d\times768}$ is initialized to make the
shift comparable in magnitude to $q$. Centering removes
common name components; no parameters are item-specific.

\textit{Attention.}
Standard attention over the $\sim2\times10^{4}$ video tokens
produces nearly uniform weights, making name-shifted queries
read similar averages. \ourscond{} therefore uses
QK-norm~\cite{henry2020qknorm,dehghani2023vit22b}:
layer normalization of queries and keys before their dot
product, with a learnable logit scale, makes attention
sensitive to name shifts (Fig.~\ref{fig:pipeline}).


\section{EXPERIMENTAL SETUP}

\subsection{Dataset and protocol}
AIxSuture~\cite{hoffmann2024} contains 314 five-minute open-suturing
videos from 157 medical students, each rated by three raters on
eight OSATS items (Table~\ref{tab:rubric_items}).
We follow the official student-grouped train/validation/test split
(220/48/46 videos; test set: 14 novice, 11 intermediate, 21 proficient)
and the official 3-class task based on rater GRS.
For Table~\ref{tab:main}, our models and same-encoder controls are
trained on all eight item ratings using the training split, with
checkpoints selected on validation and evaluated on the held-out test set.
Zero-shot experiments use separately trained models supervised on
only two items (Sec.~\ref{sec:zeroshot}).

\subsection{Compared models}
\label{sec:compared}
The published baselines are I3D~\cite{carreira2017i3d} and
Video Swin-T~\cite{liu2022videoswin} within a temporal segment
network~\cite{wang2016tsn}, pretrained on Kinetics-400 and trained
on class labels~\cite{hoffmann2024}.
All remaining models share our video encoders, clip sampling,
item ratings, and training recipe.
BCE and CE-8 perform \emph{description matching} against each
item's three level features $t_{i,1},t_{i,3},t_{i,5}$.
\textbf{BCE} (SigLIP-style~\cite{zhai2023siglip}) uses one shared
query and video vector $z$ for all items, scoring each of the
24 descriptions by $\sigma(s\cos(z,t_{i,k})+b)$ and applying
binary cross-entropy against $w_{i,k}$.
Like \ourscond{}, it has no item-specific parameters.
\textbf{CE-8} uses eight learned item queries, as in \oursitem{},
and matches each item vector $z_i$ to its three descriptions
through cosine similarity and softmax, supervised by the rater
weights $w_{i,k}$:
\begin{equation}
\begin{aligned}
\mathcal{L}_{\mathrm{CE}}
&=-\textstyle\sum_k w_{i,k}\log P_{i,k},\\
P_{i,k}
&=\operatorname{softmax}_k\big(s\cos(z_i,t_{i,k})\big).
\end{aligned}
\label{eq:ce}
\end{equation}
Since $w_i$ determines $\tau_i$~\eqref{eq:target}, CE-8 receives
the same rater information as \oursitem{}, differing only in
the objective.
\textbf{Regression} uses the item queries of \oursitem{} and
an MLP matching the size of $h_v$ to predict $\bar y_i$
with squared-error loss, replacing the rubric-language target
with a numerical score.
BCE and CE-8 use both readouts in Sec.~\ref{sec:readout},
with CE-8 probabilities defined in~\eqref{eq:ce}.

\subsection{Implementation details and metrics}
Training samples one random clip per segment with random cropping
(scale 0.8--1.0) and brightness/contrast jitter ($\pm10\%$);
inference averages predictions from three evenly spaced samples.
We train for 60 epochs using AdamW (weight decay 0.05), with learning
rates of $3\cdot10^{-5}$, $10^{-5}$, and $2\cdot10^{-4}$ for the
video encoder, text encoder, and remaining modules, respectively,
5\% warm-up, and cosine decay to 1\%.
The batch size is 8 across 4 A100 GPUs.
SIGReg uses 4096 random projections on each GPU's local batch
(2 videos), with $\lambda_v=\lambda_t=0.1$.
We report accuracy, macro-F1, class-wise F1, Pearson correlation
between predicted and rater GRS, and item MAE against mean rater scores.

\begin{table*}[t]
\caption{3-class skill classification and rubric scoring on the AIxSuture
test split (46 videos). Mean $\pm$ std over three runs; published
class-wise F1 scores have no reported std. Best in bold.}
\label{tab:main}
\centering
\setlength{\tabcolsep}{5pt}
\begin{tabular}{l c ccccc cc}
\toprule
 & Item-specific & \multicolumn{5}{c}{3-class skill classification} & \multicolumn{2}{c}{Rubric scores} \\
\cmidrule(lr){3-7}\cmidrule(lr){8-9}
Method & parameters & Acc & macro-F1 & F1-N & F1-I & F1-P & GRS $r$ & item MAE$\downarrow$ \\
\midrule
I3D~\cite{hoffmann2024} & -- & \pmv{0.746}{0.063} & \pmv{0.692}{0.029} & 0.857 & 0.403 & 0.751 & -- & -- \\
Video Swin-T~\cite{hoffmann2024} & -- & \pmv{0.732}{0.033} & \pmv{0.716}{0.043} & 0.837 & 0.538 & 0.772 & -- & -- \\
\midrule
Regression & yes & \pmv{0.768}{0.037} & \pmv{0.749}{0.047} & \pmv{0.843}{0.084} & \pmv{0.568}{0.066} & \pmv{0.837}{0.020} & \pmv{0.882}{0.010} & \pmv{0.458}{0.016} \\
CE-8, path readout & yes & \pmv{0.783}{0.031} & \pmv{0.755}{0.038} & \pmv{0.833}{0.034} & \pmv{0.581}{0.090} & \pmv{0.850}{0.024} & \pmv{0.846}{0.016} & \pmv{0.539}{0.019} \\
CE-8, prob.\ readout & yes & \pmv{0.746}{0.051} & \pmv{0.712}{0.069} & \pmv{0.798}{0.048} & \pmv{0.496}{0.139} & \pmv{0.841}{0.023} & \pmv{0.866}{0.019} & \pmv{0.461}{0.015} \\
\oursitem{} (ours) & yes & \pmv{\textbf{0.848}}{0.047} & \pmv{\textbf{0.829}}{0.056} & \pmv{\textbf{0.876}}{0.066} & \pmv{\textbf{0.703}}{0.095} & \pmv{\textbf{0.908}}{0.039} & \pmv{\textbf{0.908}}{0.010} & \pmv{0.486}{0.020} \\
\midrule
BCE, path readout & no & \pmv{0.754}{0.027} & \pmv{0.720}{0.031} & \pmv{0.769}{0.033} & \pmv{0.530}{0.067} & \pmv{0.861}{0.003} & \pmv{0.860}{0.001} & \pmv{0.466}{0.004} \\
BCE, prob.\ readout & no & \pmv{0.768}{0.037} & \pmv{0.738}{0.046} & \pmv{0.815}{0.060} & \pmv{0.548}{0.084} & \pmv{0.853}{0.011} & \pmv{0.858}{0.007} & \pmv{0.456}{0.009} \\
\ourscond{} (ours) & no & \pmv{0.790}{0.010} & \pmv{0.762}{0.016} & \pmv{0.810}{0.027} & \pmv{0.590}{0.032} & \pmv{0.885}{0.006} & \pmv{0.894}{0.012} & \pmv{\textbf{0.448}}{0.029} \\
\bottomrule
\end{tabular}
\end{table*}
\begin{table}[t]
\caption{Zero-shot scoring of JIGSAWS by the two-item models (no JIGSAWS item used in training). $\rho$: Spearman
correlation with the annotated GRS; IoU: hottest 10\% of the Grad-CAM map vs.\ instruments and suture. Mean $\pm$ std, 3 seeds.}
\label{tab:zeroshot}
\centering
\setlength{\tabcolsep}{3pt}
\resizebox{\columnwidth}{!}{%
\begin{tabular}{l cccc}
\toprule
 & \multicolumn{2}{c}{$\rho$ with the GRS} & \multicolumn{2}{c}{IoU with instruments and suture} \\
\cmidrule(lr){2-3}\cmidrule(lr){4-5}
 & Suturing & Knot Tying & Suturing & Knot Tying \\
\midrule
Random & \pmv{$-$0.001}{0.161} & \pmv{$-$0.002}{0.170} & 0.057 & 0.060 \\
Trial duration & 0.342 & 0.723 & -- & -- \\
\midrule
BCE, path readout & \pmv{0.247}{0.018} & \pmv{0.310}{0.041} & \pmv{0.024}{0.019} & \pmv{0.021}{0.008} \\
BCE, prob.\ readout & \pmv{0.319}{0.029} & \pmv{0.351}{0.167} & \pmv{0.088}{0.077} & \pmv{0.061}{0.048} \\
\ourscond{} & \pmv{0.272}{0.144} & \pmv{0.252}{0.126} & \pmv{\textbf{0.185}}{0.122} & \pmv{\textbf{0.140}}{0.101} \\
\bottomrule
\end{tabular}}
\end{table}

\begin{figure*}[t]
\centering
\includegraphics[width=0.86\textwidth]{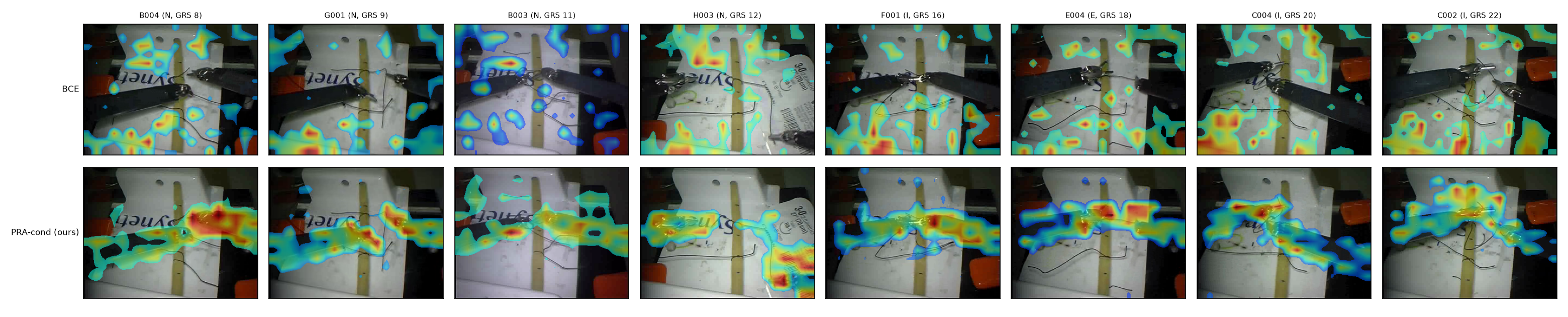}
\caption{Grad-CAM of zero-shot scoring of JIGSAWS Knot Tying by the two-item models (one
seed): map averaged over the six JIGSAWS items, hottest 25\% (for visibility; the IoU uses 10\%) overlaid on the centre frame of the fifth of eight
clips. Eight trials spread over the GRS range; N, I, E: novice, intermediate, expert (experience level). Top: BCE;
bottom: \ourscond{}.}
\label{fig:gradcam_jig}
\end{figure*}

\subsection{Zero-shot setup on JIGSAWS}\label{sec:zeroshot}
\textit{Data.} JIGSAWS~\cite{gao2014} contains bench-top da~Vinci
Suturing (39 trials) and Knot Tying (36 trials). One surgeon scores
six modified-OSATS items from 1 to 5; their sum gives GRS.
We pair the provided item names with corresponding OSATS level
descriptions (Table~\ref{tab:rubric_items}).

\textit{Training.} Using the same AIxSuture split and recipe,
BCE and \ourscond{} train only on instrument handling and knowledge
of the specific procedure (the two-item models). Neither has a
JIGSAWS counterpart, leaving all six test items unseen.
Checkpoints minimize validation MAE on these two items and transfer
without further training. \ourscond{} receives names and
descriptions; BCE receives descriptions only. Regression, CE-8,
and \oursitem{} use learned item-specific queries, preventing
transfer to unseen items.

\textit{Evaluation.} We follow AIxSuture inference, summing six
item scores into GRS and reporting task-wise Spearman's $\rho$
against annotations. Baselines use random scores or trial duration
(shorter ranks higher); models receive fixed clip counts without
duration information.

Grad-CAM~\cite{selvaraju2017gradcam} uses video-output tokens,
mean-gradient channel weights, and rectified channel summation.
It targets the closest level description through cosine similarity
for BCE or negative Euclidean distance from $h_v(z_i)$ for
\ourscond{}. Maps are averaged across items, upsampled, and
thresholded at the hottest 10\%. Grounding
DINO~\cite{liu2024groundingdino} prompts, including ``robotic
grasper'' and ``suture thread'', followed by
SAM~2.1~\cite{ravi2024sam2}, generate instrument and suture masks.
We union masks after excluding static background, bright pad,
coloured clamps, and the Knot Tying tube. Following
\cite{cheng2026surgxbench}, IoU is averaged over the centre frames
of each trial's eight clips; random 10\% pixel selection gives
the chance baseline.
\vspace{-1mm}
\section{RESULTS}
\label{sec:results}


\subsection{Main results}
In Table~\ref{tab:main}, \oursitem{} reaches
$0.848\pm0.047$ accuracy and $0.829\pm0.056$ macro-F1, 0.113 macro-F1 above Video Swin-T and 0.137 above I3D, and is the
best model in accuracy, macro-F1 and every class-wise F1. The largest gain is on the intermediate class (F1-I $0.703$ vs.\
$0.538$ for Video Swin-T and $0.403$ for I3D), the class whose item scores lie mostly between the described levels. Without item-specific parameters, \ourscond{} improves on BCE
with path readout by 0.042 macro-F1 and 0.060 F1-I, with a
macro-F1 standard deviation of 0.016 and the lowest item MAE
(0.448).


\subsection{Sensitivity to the temperature}
\label{sec:analysis}

\begin{figure}[t]
\centering
\includegraphics[width=\columnwidth]{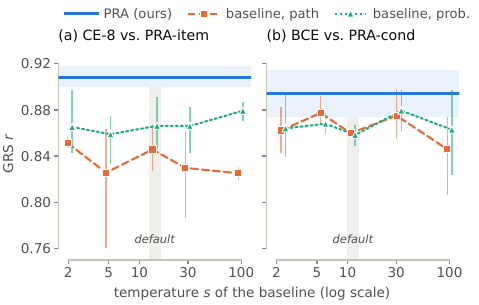}
\caption{GRS correlation on the AIxSuture test split over the temperature $s$ of the description-matching baselines
(learned value, log scale). Points: mean over 3 seeds at one $s$; bars: range over the seeds; path and probability
readouts are drawn slightly apart. Grey: default $s$ (Table~\ref{tab:main}). Blue line and band: \ours{}, which has no
temperature (mean and range over 3 seeds).}
\label{fig:temperature}
\end{figure}

BCE, CE-8 and \ours{} use the same rater weights $w_i$~\eqref{eq:weights},
but differ in their training targets. BCE and CE-8 match these weights
through similarity-based sigmoid or softmax outputs, whereas \ours{}
predicts a target feature $\tau_i$ formed from the same
weights~\eqref{eq:target}. Thus, \ours{} directly supervises the predicted
feature's position, while BCE and CE-8 supervise its similarity-based
outputs.

\textbf{\ours{} has no temperature to tune, and no tested temperature brings the baselines to its GRS correlation.}
BCE and CE-8 scale their cosine similarities by a learnable temperature $s$~\eqref{eq:ce}, which stays within 20\% of
its initial value (BCE: 10, as in SigLIP; CE-8: $1/0.07$, as in CLIP). We retrain both with $s$ initialized to 2, 5,
30 and 100 (BCE bias $b=-s$; Fig.~\ref{fig:temperature}). The best CE-8 seed at any $s$ stays below the
worst \oursitem{}, and the BCE mean stays below that of \ourscond{} at every $s$ ($0.846$--$0.879$
vs.\ $0.894$), although their seed ranges overlap.


\subsection{Zero-shot scoring on JIGSAWS}
\label{sec:zs}
\textbf{In zero-shot scoring, \ourscond{} is not the most accurate model, but it looks at the relevant visual cues.}
All models rank the trials of both tasks above random (Table~\ref{tab:zeroshot}), though below a trial-duration
baseline; \ourscond{} has lower mean correlations and greater variation
across seeds than BCE with probability readout. Its Grad-CAM maps, however,
overlap the instruments and the suture more than twice as much as those of BCE and as chance, and in
Fig.~\ref{fig:gradcam_jig} its hottest regions lie along the instruments, the thread and the knot, whereas those of BCE
spread over the pad and the background. We interpret this as an effect of the predictive objective: LeVLJEPA~\cite{kuhn2026}, which makes the text target
predictable from the video instead of maximizing their similarity, keeps markedly stronger dense features than
contrastive training, and such spatial detail is what Grad-CAM relies on.

\section{CONCLUSIONS}
In this work, we proposed Predictive Rubric Alignment (PRA), an alternative
to probability-based scoring for surgical skill assessment. \ours{} scores surgical videos by predicting a rater-weighted
combination of level-description embeddings and projecting it
onto the description path. \oursitem{} uses learned item queries,
while \ourscond{} derives queries from item names to accept new
rubrics as text. On AIxSuture, \ours{} improves classification
over published baselines and same-encoder controls, especially
for intermediate skill. Trained on only two AIxSuture items,
\ourscond{} yields positive correlations on unseen JIGSAWS items
and greater Grad-CAM overlap with instruments and suture than
description matching. Our evaluation is limited by dataset size
and rubric diversity; future work will expand training and
evaluation across surgical tasks and rubrics.



\begin{thebibliography}{99}
\bibitem{wang2016tsn} L. Wang \emph{et al.}, ``Temporal segment networks: Towards good practices for deep action
  recognition,'' in \emph{Proc. ECCV}, 2016.
\bibitem{carreira2017i3d} J. Carreira and A. Zisserman, ``Quo vadis, action recognition? A new model and the Kinetics
  dataset,'' in \emph{Proc. CVPR}, 2017.
\bibitem{liu2022videoswin} Z. Liu \emph{et al.}, ``Video Swin Transformer,'' in \emph{Proc. CVPR}, 2022.
\bibitem{zhai2023siglip} X. Zhai, B. Mustafa, A. Kolesnikov, and L. Beyer, ``Sigmoid loss for language image
  pre-training,'' in \emph{Proc. ICCV}, 2023.
\bibitem{vjepa21}
L.~Mur-Labadia et al., ``V-JEPA 2.1: Unlocking dense features
in video self-supervised learning,'' arXiv:2603.14482, 2026.
\bibitem{featuregemma}
H.~Schechter Vera et al., ``EmbeddingGemma: Powerful and lightweight
text representations,'' arXiv:2509.20354, 2025.
\bibitem{henry2020qknorm} A. Henry, P. R. Dachapally, S. S. Pawar, and Y. Chen, ``Query-key normalization for
  transformers,'' in \emph{Findings of EMNLP}, 2020.
\bibitem{dehghani2023vit22b} M. Dehghani \emph{et al.}, ``Scaling vision transformers to 22 billion parameters,'' in
  \emph{Proc. ICML}, 2023.
\bibitem{selvaraju2017gradcam} R. R. Selvaraju \emph{et al.}, ``Grad-CAM: Visual explanations from deep networks via
  gradient-based localization,'' in \emph{Proc. ICCV}, 2017.
\bibitem{liu2024groundingdino} S. Liu \emph{et al.}, ``Grounding DINO: Marrying DINO with grounded pre-training for
  open-set object detection,'' in \emph{Proc. ECCV}, 2024.
\bibitem{ravi2024sam2} N. Ravi \emph{et al.}, ``SAM 2: Segment anything in images and videos,'' arXiv:2408.00714, 2024.
\bibitem{birkmeyer2013}
J.~D. Birkmeyer et al., ``Surgical skill and complication rates after bariatric surgery,'' \emph{N. Engl. J. Med.}, vol.~369, no.~15, pp.~1434--1442, 2013.

\bibitem{martin1997}
J.~A. Martin et al., ``Objective structured assessment of technical skill (OSATS) for surgical residents,'' \emph{Br. J. Surg.}, vol.~84, no.~2, pp.~273--278, 1997.

\bibitem{igaki2023}
T.~Igaki \emph{et al.}, ``Automatic surgical skill assessment system based on concordance of standardized surgical field development using artificial intelligence,'' \emph{JAMA Surg.}, vol.~158, no.~8, Art.~no.~e231131, 2023.

\bibitem{funke2019}
I.~Funke, S.~T. Mees, J.~Weitz, and S.~Speidel, ``Video-based surgical skill assessment using 3D convolutional neural networks,'' \emph{Int. J. Comput. Assist. Radiol. Surg.}, vol.~14, no.~7, pp.~1217--1225, 2019.

\bibitem{hoffmann2024}
H.~Hoffmann et al., ``AIxSuture: Vision-based assessment of open suturing skills,'' \emph{Int. J. Comput. Assist. Radiol. Surg.}, vol.~19, no.~6, pp.~1045--1052, 2024.
\bibitem{li2022ordinalclip}
W.~Li et al., ``OrdinalCLIP: Learning rank prompts for language-guided ordinal regression,'' in \emph{Proc. NeurIPS}, 2022.

\bibitem{anastasiou2023}
D.~Anastasiou, Y.~Jin, D.~Stoyanov, and E.~Mazomenos, ``Keep your eye on the best: Contrastive regression transformer for skill assessment in robotic surgery,'' \emph{IEEE Robot. Autom. Lett.}, vol.~8, no.~3, pp.~1755--1762, 2023.
\bibitem{mukhoti2023pacl}
J.~Mukhoti et al.,
``Open vocabulary semantic segmentation with patch aligned contrastive learning,''
in \emph{Proc. CVPR}, 2023, pp.~19413--19423.
\bibitem{zhang2023liqe}
W.~Zhang, G.~Zhai, Y.~Wei, X.~Yang, and K.~Ma,
``Blind image quality assessment via vision-language correspondence: A multitask learning perspective,''
in \emph{Proc. CVPR}, 2023, pp.~14071--14081.

\bibitem{liu2025clippcqa}
Y.~Liu, Y.~Zhang, Z.~Shan, and Y.~Xu,
``CLIP-PCQA: Exploring subjective-aligned vision-language modeling for point cloud quality assessment,''
in \emph{Proc. AAAI}, 2025, pp.~5694--5702.
\bibitem{dong2023maskclip}
X.~Dong et al.,
``MaskCLIP: Masked self-distillation advances contrastive language-image pretraining,''
in \emph{Proc. CVPR}, 2023, pp.~10995--11005.
\bibitem{larson2005}
J.~L. Larson \emph{et al.}, ``Feasibility, reliability and validity of an operative performance rating system for evaluating surgery residents,'' \emph{Surgery}, vol.~138, no.~4, pp.~640--649, 2005.

\bibitem{toale2023}
C.~Toale, M.~Morris, and D.~O. Kavanagh, ``Why do residents fail simulation-based assessments of operative competence? A qualitative analysis,'' \emph{Global Surg. Educ.}, vol.~2, Art.~no.~83, 2023.

\bibitem{rau2025}
A.~Rau \emph{et al.}, ``Systematic evaluation of large vision-language models for surgical artificial intelligence,'' arXiv:2504.02799, 2025.

\bibitem{cheng2026surgxbench}
J.~Cheng \emph{et al.}, ``SurgXBench: Explainable vision-language model benchmark for surgery,'' in \emph{Proc. WACV}, 2026, pp.~8188--8198.

\bibitem{wu2024qalign}
H.~Wu et al., ``Q-Align: Teaching LMMs for visual scoring via discrete text-defined levels,'' in \emph{Proc. ICML}, 2024, pp.~54015--54029.

\bibitem{you2025deqa}
Z.~You, X.~Cai, J.~Gu, T.~Xue, and C.~Dong, ``Teaching large language models to regress accurate image quality scores using score distribution,'' in \emph{Proc. CVPR}, 2025, pp.~14483--14494.

\bibitem{kuhn2026}
L.~Kuhn, G.~Serra, R.~Balestriero, and F.~Buettner, ``LeVLJEPA: End-to-end vision-language pretraining without negatives,'' \emph{arXiv:2607.00784}, 2026.

\bibitem{gao2014}
Y.~Gao et al., ``JHU-ISI gesture and skill assessment working set (JIGSAWS): A surgical activity dataset for human motion modeling,'' in \emph{MICCAI Workshop: M2CAI}, 2014.
\bibitem{tang2020}
Y.~Tang et al., ``Uncertainty-aware score distribution learning for action quality assessment,'' in \emph{Proc. CVPR}, 2020, pp.~9839--9848.

\bibitem{xu2024}
H.~Xu et al., ``Vision-language action knowledge learning for semantic-aware action quality assessment,'' in \emph{Proc. ECCV}, 2024.

\bibitem{radford2021}
A.~Radford et al., ``Learning transferable visual models from natural language supervision,'' in \emph{Proc. ICML}, 2021, pp.~8748--8763.

\bibitem{cao2020coral}
W.~Cao, V.~Mirjalili, and S.~Raschka, ``Rank consistent ordinal regression for neural networks with application to age estimation,'' \emph{Pattern Recognit. Lett.}, vol.~140, pp.~325--331, 2020.

\bibitem{wang2023l2rclip}
R.~Wang \emph{et al.}, ``Learning-to-rank meets language: Boosting language-driven ordering alignment for ordinal classification,'' in \emph{Proc. NeurIPS}, 2023.

\bibitem{balestriero2025}
R.~Balestriero and Y.~LeCun, ``LeJEPA: Provable and scalable self-supervised learning without the heuristics,'' \emph{arXiv:2511.08544}, 2025.

\bibitem{cheng2026trajpred}
J.~Cheng et al., ``TrajPred: Trajectory-conditioned joint embedding
prediction for surgical instrument-tissue interaction recognition
in vision-language models,'' arXiv:2603.06999, 2026.
\bibitem{kowalewski2014beyond}
T. M. Kowalewski et al.,
``Beyond task time: Automated measurement augments fundamentals
of laparoscopic skills methodology,''
\emph{J. Surg. Res.}, vol. 192, no. 2, pp. 329--338, 2014,
doi: 10.1016/j.jss.2014.05.077.
\bibitem{cheng2026lavif}
J. Cheng, S. Tripathi, S. Liu, X. Yu, and S. Lin,
``LAVIFT: Latent-Action-Guided Vision Fine-Tuning for Surgical Interaction Recognition,''
\emph{arXiv preprint arXiv:2607.19889}, 2026.

\bibitem{agarwala2023temperature}
A. Agarwala, J. Pennington, Y. Dauphin, and S. Schoenholz,
``Temperature check: Theory and practice for training models with softmax-cross-entropy losses,''
\emph{Transactions on Machine Learning Research}, 2023.

\bibitem{allen2026feedback}
R. Allen, I. de Vries, T. M. M. Murray, N. Ball, N. McGuire,
S. Mann, G. B. Mizubuti, P. Szasz, and B. Zevin,
``Examining the quality and quantity of verbal feedback in the operating
room: A multi-specialty study in a Canadian context,''
\emph{Journal of Surgical Education}, 2026.

\end{thebibliography}
\end{document}